\documentclass[10pt,twocolumn,letterpaper]{article}

\usepackage[pagenumbers]{cvpr} 

\definecolor{cvprblue}{rgb}{0.21,0.49,0.74}
\usepackage[pagebackref,breaklinks,colorlinks,allcolors=cvprblue]{hyperref}
\usepackage[utf8]{inputenc}
\usepackage[table]{xcolor}
\usepackage{booktabs}
\usepackage{amssymb}
\usepackage{pifont}
\usepackage{pifont}
\newcommand{\xmark}{\ding{55}}

\usepackage{lipsum}
\usepackage{array}
\usepackage{times}
\usepackage{epsfig}
\usepackage{graphicx}
\usepackage{float}
\usepackage{wrapfig}
\usepackage{amsmath,amssymb,amsthm}
\usepackage{algorithm,algorithmicx,algpseudocode}
\usepackage{bm,xspace}
\usepackage{comment}
\usepackage{multirow}
\usepackage{balance}
\usepackage{url}
\usepackage{booktabs}
\usepackage{etoolbox,siunitx}
\usepackage{calc}
\usepackage{pifont,hologo}
\usepackage{color}
\usepackage{adjustbox}
\usepackage{amsmath}
\usepackage{enumitem}
\usepackage{bbding}  %
\usepackage[normalem]{ulem}  %
\usepackage{contour}
\PassOptionsToPackage{square,numbers,comma,sort}{natbib}

\definecolor{blue}{HTML}{004bb3}
\definecolor{red}{HTML}{cc1100}
\definecolor{orange}{HTML}{cc7700}
\definecolor{gray}{HTML}{efefef}
\definecolor{darkgreen}{HTML}{228B22}
\definecolor{darkgray}{HTML}{757575}

\definecolor{cite}{HTML}{3270b5}
\definecolor{link}{HTML}{b53532}
\definecolor{link}{HTML}{cc1100}
\definecolor{scratch}{HTML}{001219}
\definecolor{pretrain}{HTML}{0A9396}

\contourlength{0.8pt}

\renewcommand{\eqref}[1]{Eq.~\ref{#1}}

\newcolumntype{x}[1]{>{\centering\arraybackslash}p{#1}}
\newcolumntype{y}[1]{>{\raggedright\arraybackslash}p{#1}}
\newcolumntype{z}[1]{>{\raggedleft\arraybackslash}p{#1}}

\DeclareMathSymbol{@}{\mathord}{letters}{"3B}

\makeatletter
\DeclareRobustCommand\onedot{\futurelet\@let@token\@onedot}
\def\@onedot{\ifx\@let@token.\else.\null\fi\xspace}

\newcommand*{\Rom}[1]{\expandafter\@slowromancap\romannumeral #1@}
\newcommand*{\rom}[1]{\expandafter\romannumeral #1}

\def\1{\bm{1}}

\DeclareMathAlphabet{\mathsfit}{\encodingdefault}{\sfdefault}{m}{sl}
\SetMathAlphabet{\mathsfit}{bold}{\encodingdefault}{\sfdefault}{bx}{n}

\let\originalleft\left
\let\originalright\right
\renewcommand{\left}{\mathopen{}\mathclose\bgroup\originalleft}
\renewcommand{\right}{\aftergroup\egroup\originalright}

\def\paperID{5252} 
\def\confName{CVPR}
\def\confYear{2026}

\title{Depth Any Panoramas: A Foundation Model for Panoramic Depth Estimation}

\author{%
\vspace{0.5em}%
Xin Lin\textsuperscript{2}\quad
Meixi Song\textsuperscript{1}\quad 
Dizhe Zhang\textsuperscript{1}\footnotemark[2] \footnotemark[3]\quad
Wenxuan Lu\textsuperscript{1}\quad
Haodong Li\textsuperscript{2}\quad \\
Bo Du\textsuperscript{3}\quad
Ming-Hsuan Yang\textsuperscript{4}\quad
Truong Nguyen\textsuperscript{2}\footnotemark[1]\quad
Lu Qi\textsuperscript{1,3}\footnotemark[1] \footnotemark[2] \\
\vspace{0.5em}\small
\textsuperscript{1} Insta360 Research\quad
\textsuperscript{2} University of California, San Diego\quad
\textsuperscript{3} Wuhan University\quad
\textsuperscript{4} University of California, Merced\quad
\\
}
\begin{document}

\twocolumn[{%
\renewcommand\twocolumn[1][]{#1}%
\vspace{-1em}
\maketitle
\centering
\vspace{-1em}
\includegraphics[width=0.99\linewidth]{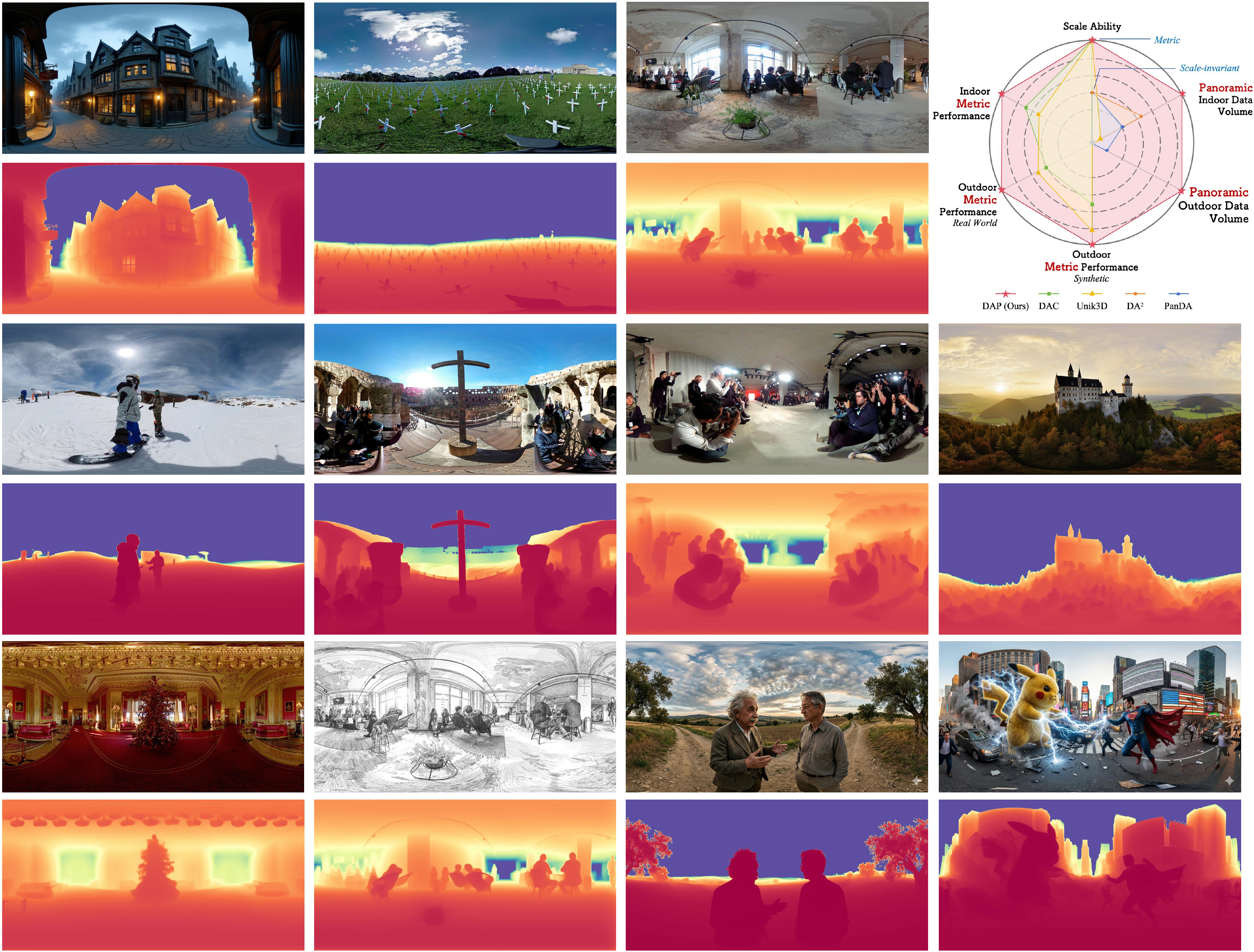}
\vspace{-1em}
\captionof{figure}{Metric depth visualizations generated by DAP from diverse panoramic inputs. 
For clarity, each depth map is displayed using its own adaptive truncation range. DAP achieves robust, metrically consistent panoramic depth across diverse real-world scenes, highlighting the power of large-scale data and model designing.
\vspace{1em}
}
\label{teaser}
}]

\let\thefootnote\relax\footnotetext{
$^*$ Equal advising \hspace{5pt}
$^\dagger$ Corresponding author \hspace{5pt}
$^\ddagger$ Project leader
}

\maketitle
\begin{abstract}
In this work, we present a panoramic metric depth foundation model that generalizes across diverse scene distances. We explore a data-in-the-loop paradigm from the view of both data construction and framework design. We collect a large-scale dataset by combining public datasets, high-quality synthetic data from our UE5 simulator and text-to-image models, and real panoramic images from the web. 
To reduce domain gaps between indoor/outdoor and synthetic/real data, we introduce a three-stage pseudo-label curation pipeline to generate reliable ground truth for unlabeled images. 
For the model, we adopt DINOv3-Large as the backbone for its strong pre-trained generalization, and introduce a plug-and-play range mask head, sharpness-centric optimization, and geometry-centric optimization to improve robustness to varying distances and enforce geometric consistency across views. Experiments on multiple benchmarks (e.g., Stanford2D3D, Matterport3D, and Deep360) demonstrate strong performance and zero-shot generalization, with particularly robust and stable metric predictions in diverse real-world scenes. The project page can be found at: \href{https://insta360-research-team.github.io/DAP_website/}
{https://insta360-research-team.github.io/DAP\_website/}

\end{abstract}   
\section{Introduction}
\label{sec:intro}

\begin{table*}[t!]
\caption{Comparison of training data compositions used by recent panoramic depth estimation methods. 
Unlike previous approaches, which rely on limited or domain-specific datasets, our DAP data engine scales up to 2M panoramas across both indoor/outdoor and synthetic/real domains, providing a unified and comprehensive data foundation for panoramic depth modeling. * in DA$^{2}$ refers to pseudo-panoramic data generated from perspective images through P2E projection and out-painting model.}
\begin{center}{
\setlength{\tabcolsep}{3mm}{
\begin{tabular}{lccccccc}
\toprule
\multirow{2}{*}{Method} & \multirow{2}{*}{Metric Ability} & \multirow{2}{*}{Dataset Type} & \multirow{2}{*}{Number} & \multicolumn{2}{c}{Scene} & \multicolumn{2}{c}{Panorama} \\
& & & & Indoor & Outdoor & Synthetic & Real World \\
\midrule
PanDA~\cite{cao2025panda}& \ding{55} & Panorama & 122k & \checkmark & \checkmark & \checkmark (20k) & \checkmark (102k) \\
DA$^{2}$~\cite{da2}& \ding{55} & Mixed* & 606k & \checkmark & \checkmark & \checkmark (63k) & \ding{55} \\
Unik3D~\cite{piccinelli2025unik3d}& \checkmark & Mixed & 694k & \checkmark & \checkmark & \checkmark (1k) & \ding{55} \\
DAC~\cite{DepthAnyCamera}& \checkmark & Mixed & 800k & \checkmark (670k) & \checkmark (130k) & \ding{55} & \ding{55} \\
DAP (Ours) & \checkmark & Panorama & 2M & \checkmark (500k) & \checkmark (1.5M) & \checkmark (300k) & \checkmark (1.7M) \\

\bottomrule
\end{tabular}}}
\label{table:panodata}
\end{center}
\end{table*}

The estimation of panoramic depth has attracted increasing attention due to the full 360\textdegree{} $\times$ 180\textdegree{} coverage of the surrounding environment for spatial intelligence. It has benefited various robotic applications, such as omnidirectional obstacle avoidance during navigation tasks.

Despite its importance, the panoramic depth estimation still lags behind. Both panorama-specific relative/scale-invariant methods (e.g., Panda~\cite{cao2025panda}, Depth Anywhere~\cite{wang2024depthanywhere}, DA$^2$~\cite{da2}) and unified metric-depth (DAC~\cite{DepthAnyCamera}, Unik3D~\cite{piccinelli2025unik3d}) frameworks struggle to generalize to diverse real-world scenes, particularly outdoors. A possible reason is the limited scale and diversity of existing data, due to the high cost of data collection and annotation.

Motivated by this, we explore the data-in-the-loop paradigm for panoramic depth estimation, which raises two key challenges for data scaling: constructing large-scale datasets with reliable, high-quality ground truth, and designing models that can effectively adapt to such data scaling. Addressing both challenges is essential for a geometry-consistent and generalizable panoramic foundation model.

At the data level, we integrate the indoor dataset Structured3D~\cite{zheng2020structured3d}, synthesize 90K high-quality outdoor samples using UE5-based AirSim360 simulator~\cite{simulator}, and collect about 1.7M unlabeled panoramic images from internet and generate 200k indoor panoramas by DiT360~\cite{feng2025dit360}. To reduce performance degradation caused by domain gaps between indoor/outdoor and synthetic/real data, we propose a two-stage pseudo-label curation pipeline. 
In the first stage, we train a Scene-Invariant Labeler on a balanced mix of Structured3D and AirSim360, then generate pseudo labels for the 1.9M unlabeled images. 
In the second stage, we select the top 60k highest-confidence pseudo-labeled samples in both indoor and outdoor scenes using a discriminator, and train the Realism-Invariant Labeler on this expanded dataset to refine the pseudo labels.
Finally, in the third stage, our foundation model DAP is trained jointly on all labeled data and the refined pseudo-labeled data produced by the Realism-Invariant Labeler.

For the foundation model design, we adopt metric depth estimation for its scalability to panoramas across arbitrary distances. We use DINOv3-Large as the encoder to leverage a strong pre-trained visual priors and conventional depth head module for dense estimation~\cite{yang2024depthanythingv2}. We further introduce a set of losses to enforce robustness to distances and geometric consistency across viewpoints. Specifically, a lightweight depth-range mask head is presented to mitigate uneven depth distributions by filtering regions using distance thresholds (e.g., 10/20/50/100 m) in a plug-and-play manner. Moreover, Silog, DF-Gram, Gradient, Normal, and Point-Cloud losses are designed by a distortion-aware map to compensate for non-uniform pixel geometry in equirectangular projection.

The extensive experiments on both indoor and outdoor test sets (e.g., Stanford2D3D, Matterport3D, and Deep360) demonstrate the strong generalization and effectiveness of our foundation model across diverse benchmarks. 
Beyond quantitative superiority, as shown in Figure \ref{teaser}, our model exhibits excellent visual consistency and scale-awareness, producing realistic depth predictions in challenging real-world scenarios with complex geometry, distant regions, and sky areas. These results highlight the model’s robustness and strong adaptability to both synthetic and real-world environments.
In summary, the main contributions can be summarized as follows:

\begin{itemize}

    \item We construct a large-scale panoramic dataset with more than 2M samples across synthetic and real domains. It includes high-quality indoor data from Structured3D, 90K photorealistic outdoor panoramas rendered with the UE5-based AirSim360 simulator, and 1.9M filtered panoramas collected from internet and DiT360, enabling diverse and scalable depth supervision.

    \item We design a three-stage pipeline that progressively refines pseudo labels and bridges both the synthetic–real and indoor–outdoor domain gaps. This is achieved through multiple curation techniques and large-scale semi-supervised learning to enhance cross-domain generalization.

    \item We incorporate a plug-and-play range mask together with geometry- and sharpness-oriented optimization, where the associated loss terms ($\mathcal{L}_{SILog}$, $\mathcal{L}_{DF}$, $\mathcal{L}_{grad}$, $\mathcal{L}_{normal}$, and $\mathcal{L}_{pts}$), ensuring metric consistency and structural fidelity across diverse panoramas.

    \item Comprehensive evaluations show that our model generalizes well to open real-world scenarios, producing scale-consistent and perceptually coherent depth maps. It achieves state-of-the-art performance both quantitatively on synthetic benchmarks and qualitatively on diverse real-world panoramas.
    
\end{itemize}

\section{Related work}
\label{sec:related}

\noindent \textbf{Perspective Depth Estimation.}
With the rapid progress of deep learning and large-scale perspective depth datasets, perspective depth estimation has advanced rapidly, with recent metric and scale-invariant models achieving strong performance,
such as UniDepth~\citep{piccinelli2024unidepth,piccinelli2025unidepthv2}, Metric3D~\citep{hu2024metric3d,yin2023metric3d}, DepthPro~\citep{bochkovskiydepth}, and MoGe~\citep{wang2025moge,wang2025moge2}.
Some relative depth estimation methods have greatly benefited from data scaling, with models like DepthAnything~\citep{yang2024depthanything,yang2024depthanythingv2} showing impressive zero-shot generalization.
Recently, some methods fine-tune large pre-trained generative models with strong prior capabilities, such as Stable Diffusion~\citep{rombach2022high,ho2020denoising} and FLUX~\citep{flux2024}, on limited but high-quality datasets, achieving competitive results~\citep{marigold,he2024lotus,wang2025jasmine,li2025stereodiff}.
Nevertheless, the perspective paradigm inherently restricts perception to a limited field of view (FoV), failing to capture the complete 360\textdegree{} spatial geometry of a scene.

\noindent \textbf{Panoramic Depth Estimation.} \noindent\textit{In-domain.}~Early methods focus on in-domain settings, where models are trained and evaluated on the same dataset.
To address the severe distortions of the equirectangular projection (ERP), there are two main directions: distortion-aware designs~\cite{tateno2018distortion, zhuang2022acdnet, shen2022panoformer, yun2023egformer, mohadikar2025omnidiffusion, lee2025hush} and projection-driven strategies~\cite{wang2020bifuse, jiang2021unifuse, peng2023high, bai2024glpanodepth, ai2023hrdfuse, ai2024elite360d, ai2024elite360m, li2022omnifusion, rey2022360monodepth, benny2025sphereuformer, shen2024revisiting, deng2025omnistereo}.  
However, over-reliance on in-domain training often leads to overfitting and limited generalization, motivating recent efforts toward zero-shot and cross-domain panoramic depth estimation.

\begin{figure*}[t]
\centering
\includegraphics[width=1\textwidth]{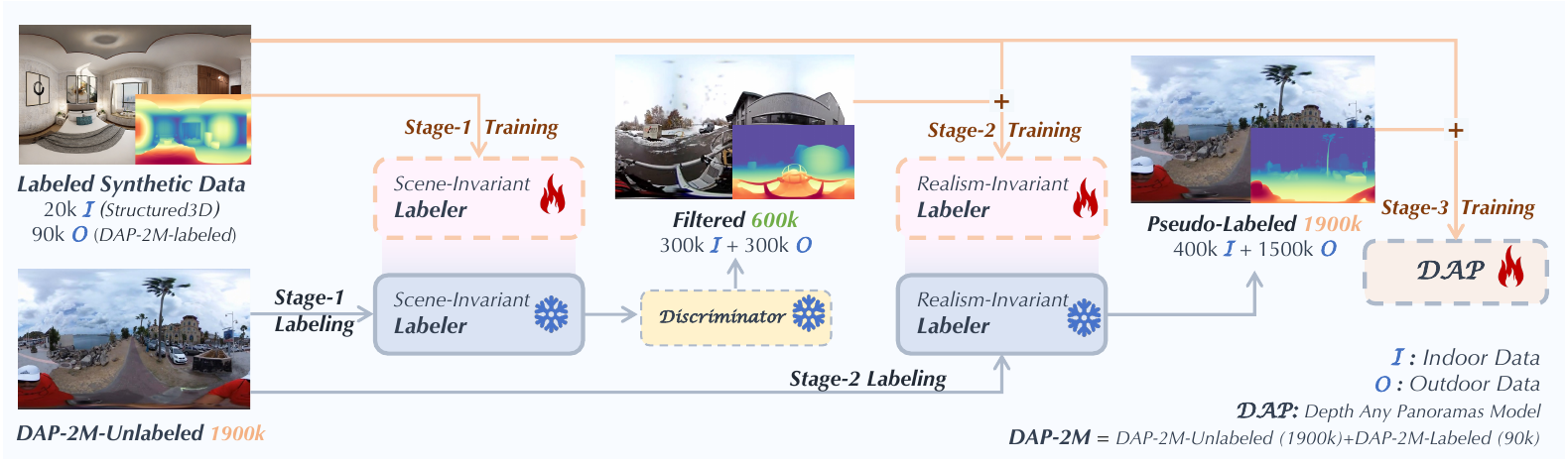}
\caption{Overview of the proposed progressive three-stage pipeline. Stage~1 trains a \textbf{Scene-Invariant Labeler} on high-quality synthetic indoor and outdoor data to provide strong initialization. 
Stage~2 introduces a \textbf{Realism-Invariant Labeler}, where a PatchGAN-based discriminator selects 300K indoor and 300K outdoor high-confidence pseudo-labeled samples to mitigate domain gaps between synthetic and real data. 
Stage~3 performs \textbf{DAP} training on all labeled and pseudo-labeled data, enabling large-scale semi-supervised learning and strong generalization across real-world panoramic scenes.
}
\label{pipeline}
\end{figure*}

\noindent\textit{Zero-shot.} Compared with in-domain training, zero-shot panoramic depth estimation is more practical for cross-domain applications due to its stronger generalization ability. Existing approaches can be grouped into three categories:
The first leverages pretrained perspective depth models to generate pseudo labels for panoramas, as in Depth Anywhere~\cite{wang2024depth} and PanDA~\cite{cao2025panda}, which distill knowledge via cube projection or semi-supervised learning on large-scale unlabeled panoramas.
More recently, DA$^2$~\cite{da2} improves zero-shot performance by expanding training data through perspective-to-ERP conversion and diffusion-based out-painting, combined with a distortion-aware transformer. A third line of work aims at universal, metric-capable camera modeling: Depth Any Camera~\cite{DepthAnyCamera} unifies diverse imaging geometries under an ERP representation with geometry-driven augmentation, while UniK3D~\cite{piccinelli2025unik3d} reformulates depth in spherical coordinates using harmonic ray representations to enhance wide-FoV generalization.

\section{Method}

We introduce a scalable panoramic depth foundation model that unifies metric estimation across diverse domains and scene types. 
Our approach consists of three main components: 
a large-scale data engine for data scaling across synthetic and real panoramas (Sec.~\ref{method:data}), 
a three-stage pipeline for effectively exploiting large-scale unlabeled data (Sec.~\ref{method:framework}), 
and a geometry-consistent network design with multi-loss optimization for high-quality metric depth estimation (Sec.~\ref{method:design}).

\subsection{Data Engine}
\label{method:data}

\noindent \textbf{Overview.} To enable the scaling of the panoramic depth estimation, we construct a comprehensive data engine that unifies diverse sources in synthetic and real domains, as summarized in Table~\ref{table:panodata}. 
Compared to the datasets used in previous panoramic depth methods such as PanDA \cite{cao2025panda}, UniK3D \cite{piccinelli2025unik3d}, and DAC \cite{DepthAnyCamera}, our data engine achieves the largest panorama scale with about 2 million samples and the broadest domain coverage, spanning indoor/outdoor and synthetic/real-world panoramas. 
It serves as the cornerstone for building our foundation model by supporting both large-scale supervised and semi-supervised training.

\noindent \textbf{Simulated Outdoor Scene Data.} To address the scarcity of outdoor panoramic supervision, we construct a synthetic outdoor dataset named DAP-2M-Labeled using the high-fidelity simulation platform Airsim360~\cite{simulator}.
We simulate drone flights following mid- and low-altitude trajectories across diverse environments to capture panoramic imagery and corresponding depth maps under realistic illumination and environmental conditions.
In total, over 90K panoramic frames with pixel-aligned depth annotations are collected from five representative outdoor scenes, including \textit{New York City}, \textit{SF City}, \textit{Downtown West}, \textit{City Park}, and \textit{Rome}, covering more than \text{26,600 complete panoramic sequences}.

\noindent \textbf{Unlabeled Data,} We have collected 250K panoramic videos from the internet and processed them into image frames.
After a careful curation procedure and filtering out samples with unreasonable horizons, we obtain a total of 1.7 million high-quality panoramic images. 
We then employ the large multimodal model Qwen2-VL \cite{wang2024qwen2} to automatically categorize these panoramas into indoor and outdoor scenes, yielding approximately \text{250K indoor} and \text{1.45M outdoor} samples.
Since indoor panoramas are relatively scarce in real-world data, we further supplement this domain by generating an additional \text{200K indoor samples} using the state-of-the-art panoramic generation model DiT-360~\cite{feng2025dit360}.
Together, these collections form our DAP-2M-Unlabeled, which provides abundant coverage of diverse environments for pretraining and semi-supervised learning.

\subsection{Three-Stage Pipeline}
\label{method:framework}
As shown in Figure~\ref{pipeline}, we adopt a three-stage pipeline to efficiently exploit unlabeled large-scale data in the real world, while enhancing the learning capacity of the network through geometric and dense-detail supervision paradigms.
Our pipeline consists of three stages:  

\noindent \textbf{Stage 1. Scene-Invariant Labeler Training.}  
We first train a Scene-Invariant Labeler on 20k synthetic indoor and 90k synthetic outdoor datasets with accurate metric depth annotations.
The goal of this stage is to learn a labeling model that generalizes across both indoor and outdoor environments rather than overfitting to specific scene layouts or lighting conditions.
Training on geometrically and photometrically diverse synthetic scenes enables the labeler to learn consistent, physically grounded depth cues that generalize across domains, providing a robust initialization for generating reliable pseudo-depth labels on real-world panoramic data.

\begin{figure*}[t]
\centering
\includegraphics[width=1\textwidth]{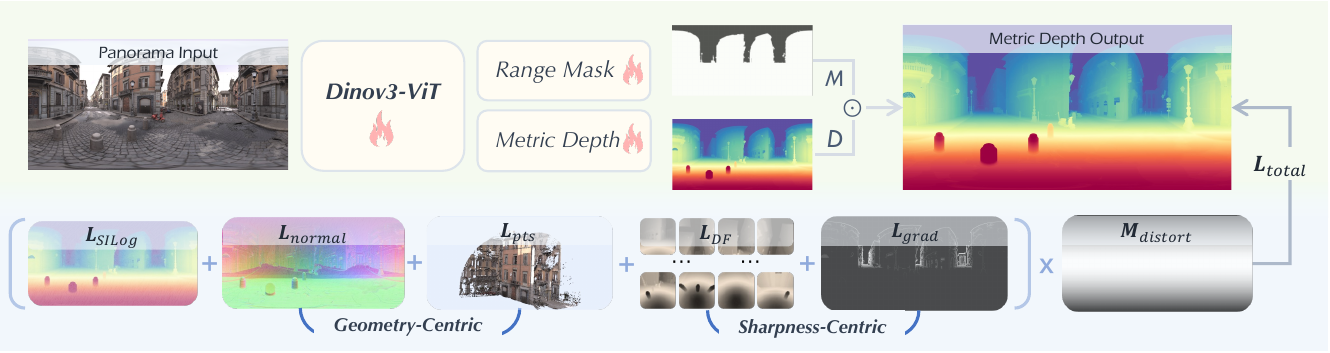}
\caption{Architecture of the proposed \textbf{DAP network}. 
Built upon DINOv3-Large~\cite{simeoni2025dinov3} as the visual backbone, our model adopts a distortion-aware depth decoder and a plug-and-play range mask head for adaptive distance control across diverse scenes. 
Training is guided by multi-level geometric and sharpness-aware losses, including $\mathcal{L}_{SILog}$, $\mathcal{L}_{DF}$, $\mathcal{L}_{grad}$, $\mathcal{L}_{normal}$, and $\mathcal{L}_{pts}$ losses, ensuring metric accuracy, edge fidelity, and geometric consistency in panoramic depth estimation.}
\label{network}
\end{figure*}

\noindent \textbf{Stage 2. Realism-Invariant Labeler Training.} We first pre-train a depth quality discriminator to assess the reliability of depth predictions, where synthetic ground-truth depth maps are treated as real samples and the Scene-Invariant Labeler outputs as fake ones, this step enables the discriminator to learn a scene-agnostic quality prior for subsequent filtering. 
For more details, please refer to the supplementary material. 
Next, we apply the Scene-Invariant Labeler to all unlabeled real images and estimate depth quality using the trained discriminator.
The top-ranked 300K indoor and 300K outdoor samples are selected as high-confidence pseudo-labeled data, which are combined with the synthetic datasets from Stage~1 to train a Realism-Invariant Labeler.
By learning from reliable pseudo labels across diverse real domains, this labeler becomes robust to appearance variations and realism-specific differences, enabling it to generalize beyond synthetic textures and lighting conditions.

\noindent \textbf{Stage 3. DAP Training.}
As shown in Table~\ref{tab:dataset}, our Depth Any Panorama (DAP) model is trained on all 1.9M pseudo-labeled data generated by the Realism-Invariant Labeler and previous labeled data. 
This enables the DAP model to effectively benefit from dense features and large-scale pseudo supervision, leading to improved generalization on real-world panoramic depth estimation.  

\begin{table}[t!]
\centering
\caption{Overview of datasets used for training DAP, covering synthetic and real, labeled and unlabeled panoramic data.}
\setlength{\tabcolsep}{4pt} 
\begin{tabular}{lcccc}
\toprule
Datasets & Indoor & Outdoor & Label & Samples \\ 
\midrule
Structured3D~\cite{zheng2020structured3d} & \ding{51} &  & \ding{51} & 18,298 \\
DAP-2M-Labeled &  & \ding{51} & \ding{51} & 90k \\
DAP-2M-Unlabeled & \ding{51} & \ding{51} & \ding{55} & 1.9M \\
\bottomrule
\end{tabular}
\label{tab:dataset}
\end{table}

\subsection{Model Design}
\label{method:design}

As illustrated in Figure~\ref{network}, our DAP network takes a panoramic image as input and uses DINOv3-Large~\cite{simeoni2025dinov3} as a visual backbone for powerful feature extraction.
We introduce two task-specific heads: a range mask head and a metric depth head. 
The range mask head outputs a binary mask $M$ that defines valid spatial regions under different distance thresholds, while the depth head predicts a dense metric depth map $D$.
To accommodate diverse environments from confined indoor spaces to large-scale outdoor scenes, we provide four plug-and-play range mask heads with distance thresholds of 10 m, 20 m, 50 m, and 100 m, allowing flexible adaptation to different spatial scales.
Each range mask head is independently optimized with a combination of weighted BCE and Dice losses:
\begin{equation}
\mathcal{L}_{mask} = 
\left\| M - M_{gt} \right\|^{2}
+ 0.5 \, \mathcal{L}_{Dice}(M, M_{gt}),
\end{equation}
where $M$ and $M_{gt}$ denote the predicted and ground-truth range masks, respectively.
The final metric depth output is obtained through element-wise multiplication of $M$ and $D$, ensuring that the predictions remain physically valid and scale-consistent in varying depth ranges.
This dual-head design enables DAP to robustly adapt to diverse scene geometries and maintain high-quality metric depth estimation across a wide spectrum of spatial conditions.
For optimization, besides adopting the SILog loss $\mathcal{L}_{SILog}$ as in previous works~\cite{cao2025panda}, we introduce a set of complementary loss functions to enhance geometric consistency and edge fidelity.

\begin{figure*}[t]
\centering
\includegraphics[width=1\textwidth]{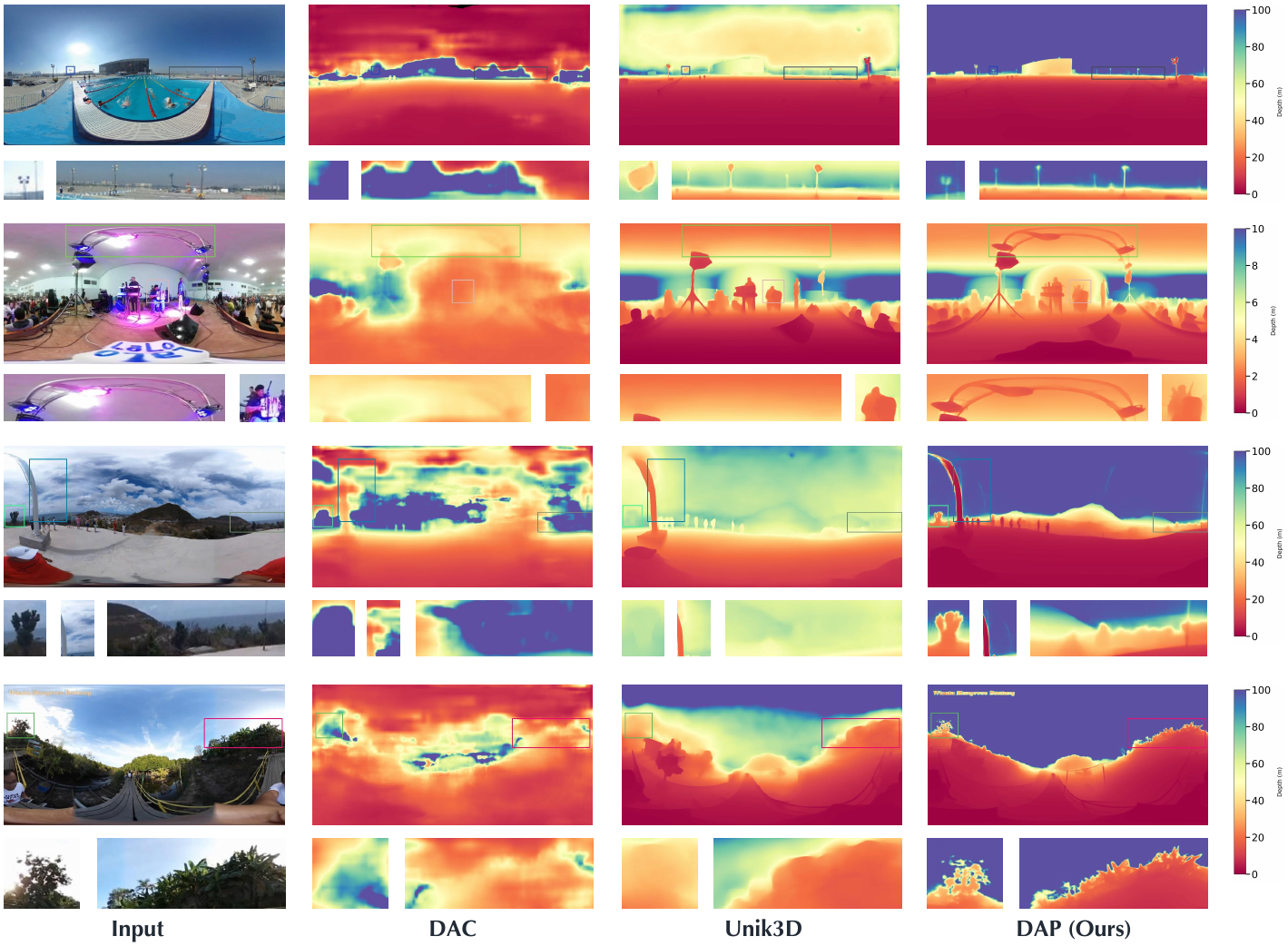}
\caption{Qualitative comparison across diverse real-world indoor and outdoor scenes. 
Our DAP produces sharper object boundaries, smoother global geometry, and superior robustness in distant and sky regions compared to DAC \cite{DepthAnyCamera} and Unik3D \cite{piccinelli2025unik3d}.}
\label{fig:qualitative}
\end{figure*}

\noindent \textbf{Sharpness-centric Optimization}
Inspired by \cite{wang2025moge, simeoni2025dinov3}, we introduce a dense fidelity constraint, termed $\mathcal{L}_{DF}$, to enhance local sharpness and structural consistency.
To mitigate geometric distortion in ERP, we first decompose each depth map into 12 perspective patches using virtual cameras positioned at the vertices of an icosahedron.
These perspective views preserve fine-grained details and avoid the stretching artifacts near poles, thus providing higher-fidelity supervision.
For each view, we apply a valid mask, normalize depth values, and compute a Gram-based similarity between the predicted and ground-truth depth maps.
The final $\mathcal{L}_{DF}$ is defined as the average loss across all $N=12$ views:
\begin{equation}
\mathcal{L}_{DF} = 
\frac{1}{N} \sum_{k=1}^{N} 
\left\|
D^{(k)}_{pred} \odot {D^{(k)}_{pred}}^{\top} 
- 
D^{(k)}_{gt} \odot {D^{(k)}_{gt}}^{\top} 
\right\|_{F}^{2},
\end{equation}
where $N=12$ denotes the number of views.

\noindent \textbf{Sharpness-oriented Gradient Refinement.}
To further enhance the sharpness of object boundaries, we introduce a gradient-based loss $\mathcal{L}_{grad}$ that explicitly focuses on high-frequency edge regions in the ERP domain.
While the $\mathcal{L}_{DF}$ strengthens dense fidelity on distortion-free perspective patches, $\mathcal{L}_{grad}$ complements it by preserving local discontinuities directly in the ERP representation.
Specifically, we compute gradient magnitude maps using Sobel operators along both $x$ and $y$ directions and derive an edge mask $M_{E}$ by thresholding the ground-truth Sobel gradient magnitudes.
The gradient loss is then applied only within these masked regions using SILog loss~\cite{eigen2014depth}:
\begin{equation}
\mathcal{L}_{grad} = 
\mathcal{L}_{SILog}(M_{E} \odot D_{pred}, \, M_{E} \odot D_{gt}).
\end{equation}
This design improves the consistency and sharpness of depth boundaries, effectively complementing the dense fidelity constraint for recovering fine geometric details.

\noindent \textbf{Geometry-centric optimization}
To improve geometric consistency, we incorporate a normal loss $\mathcal{L}_{normal}$ \cite{da2, lee2025hush}. 
Both predicted and ground-truth depth maps are converted into surface normal fields $\mathbf{n}_{pred}, \mathbf{n}_{gt} \in \mathbb{R}^{H \times W \times 3}$.
The $\mathcal{L}_{normal}$ is then defined as the L1 distance between predicted and ground-truth normals:
\begin{equation}
\mathcal{L}_{normal} =  
\lVert \mathbf{n}_{pred}(i,j) - \mathbf{n}_{gt}(i,j) \rVert_1.
\end{equation}
We further use a point cloud loss $\mathcal{L}_{pts}$. 
The depth maps are projected onto the spherical coordinate system to obtain 3D point clouds $\mathbf{P}_{pred}, \mathbf{P}_{gt} \in \mathbb{R}^{H \times W \times 3}$, and the loss is defined as:
\begin{equation}
\mathcal{L}_{pts} = 
\lVert \mathbf{P}_{pred}(i,j) - \mathbf{P}_{gt}(i,j) \rVert_1.
\end{equation}

\noindent \textbf{Overall Objective} The overall training objective is a weighted combination of all the above losses with distortion map to address the non-uniform pixel distribution \cite{lin2025one} in equirectangular projections.
The distortion map compensates for the over-representation of pixels near the poles, ensuring balanced gradient contributions across the spherical domain:
\begin{align}
\mathcal{L}_{total} 
&= M_{distort} \odot 
\big(
\lambda_{1}\mathcal{L}_{SILog} 
+ \lambda_{2}\mathcal{L}_{DF} + \lambda_{3}\mathcal{L}_{grad}  \nonumber \\
& + \lambda_{4}\mathcal{L}_{normal} 
+ \lambda_{5}\mathcal{L}_{pts} + \lambda_{6}\mathcal{L}_{mask}
\big),
\end{align}
where $\lambda_i, \; i \in \{1,2,3,4,5,6\}$, denotes a weight parameters. 

\begin{figure*}[t]
\centering
\includegraphics[width=1\textwidth]{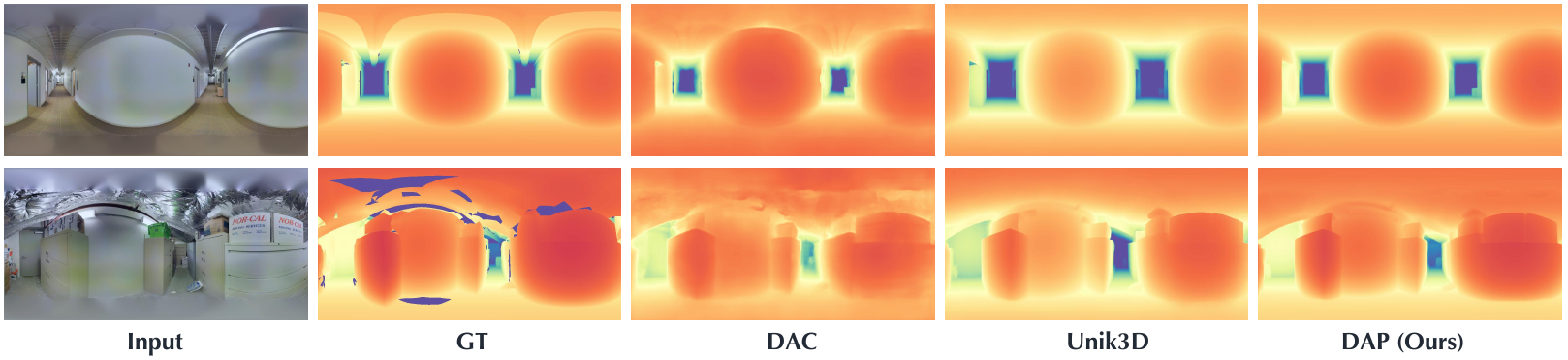}
\caption{Qualitative comparison on Stanford2D3D. Our method preserves fine structural details and demonstrates superior scale-awareness.
}
\label{fig:qualitative2}
\end{figure*}

\begin{table*}[t!]
\centering
\caption{\label{table:zero} Zero-shot comparison of panoramic metric depth estimation on three benchmarks. The \colorbox{red!25}{\textbf{best}} and \colorbox{orange!25}{\textbf{second best}} metric depth performances are highlighted. Our DAP consistently achieves the best results across all datasets, demonstrating strong generalization without fine-tuning. We also include several scale-invariant methods only for reference from DA$^2$ \cite{da2}.}
\adjustbox{max width=1\textwidth}{%
\begin{tabular}{c l| c c c | c c c | c c c }
\toprule
\multicolumn{2}{c}{\multirow{2}{*}[-0.5ex]{Methods}} & \multicolumn{3}{c}{Stanford2D3D (Indoor)} & \multicolumn{3}{c}{Matterport3D (Indoor)} & \multicolumn{3}{c}{Deep360 (Outdoor)} \\ \cmidrule(lr){3-5}\cmidrule(lr){6-8}\cmidrule(lr){9-11}
\multicolumn{2}{c}{} & \multicolumn{1}{c}{AbsRel($\downarrow$)} & \multicolumn{1}{c}{RMSE($\downarrow$)} & \multicolumn{1}{c}{$\delta_1$($\uparrow$)}
& \multicolumn{1}{c}{AbsRel($\downarrow$)} & \multicolumn{1}{c}{RMSE($\downarrow$)} & \multicolumn{1}{c}{$\delta_1$($\uparrow$)} 
& \multicolumn{1}{c}{AbsRel($\downarrow$)} & \multicolumn{1}{c}{RMSE($\downarrow$)} & \multicolumn{1}{c}{$\delta_1$($\uparrow$)} \\ \midrule

\multirow{5}{*}{\centering\rotatebox{90}{Scale-invariant}} & MoGe \cite{wang2025moge} & 0.1581 & 0.2576 & 0.7902 & 0.1004 & 0.3591 & 0.9080 & - & - & - \\ 
& VGGT \cite{wang2025vggt} & 0.1870 & 0.3350 & 0.7408 & 0.1078 & 0.3880 & 0.8870 & - & - & - \\ 
& DepthAnythingv2 \cite{yang2024depthanythingv2} & 0.3679 & 0.4339 & 0.4766 & 0.2585 & 0.7067 & 0.5842 & - & - & - \\ 
& PanDA \cite{cao2025panda} & 0.1648 & 0.2364 & 0.7326 & 0.0888 & 0.3325 & 0.9209 & - & - & - \\ 
& DA$^2$ \cite{da2} & 0.0723 & 0.1400 & 0.9545 & 0.0667 & 0.2882 & 0.9561 & - & - & - \\ 
\midrule
\multirow{3}{*}{\centering\rotatebox{90}{Metric}} & 
Unik3D~\cite{piccinelli2025unik3d} & 0.1795 & 0.4850 & 0.7823
& 0.2224 & \cellcolor{red!25}0.6680 & 0.6634 & \cellcolor{orange!25}0.0885 & \cellcolor{orange!25}6.148 & \cellcolor{orange!25}0.9293 \\ 
& DAC~\cite{DepthAnyCamera} & \cellcolor{orange!25}0.1366 & \cellcolor{orange!25}0.4509 & \cellcolor{orange!25}0.8393 & \cellcolor{orange!25}0.1803 & 0.9390 & \cellcolor{orange!25}0.7203 & 0.2611 & 8.371 & 0.6311\\
& DAP (Ours) & \cellcolor{red!25}0.0921 & \cellcolor{red!25} 0.3820 & \cellcolor{red!25}0.9135 & \cellcolor{red!25}0.1186 & \cellcolor{orange!25}0.7510 & \cellcolor{red!25}0.8518 & \cellcolor{red!25}0.0659 & \cellcolor{red!25}5.224 & \cellcolor{red!25}0.9525 \\ 
 \bottomrule
\end{tabular}}
\end{table*}

\label{sec:method}

\section{Experiment}
\label{sec:experiment}
\noindent \textbf{Training Datasets.} 
As discussed in Sec.~\ref{method:data}, the datasets used for training are summarized in Table~\ref{tab:dataset}. 
We utilize both labeled and unlabeled data from diverse indoor and outdoor environments to enhance the generalization ability of our model. 
Structured3D~\cite{zheng2020structured3d} provides high-quality synthetic indoor scenes with ground-truth depth supervision. 
DAP-2M-Labeled contains 90K real-world samples with depth annotations, while DAP-2M-Unlabeled includes 1.9M real images without depth labels, which are leveraged for pseudo-label learning in our pipeline.
For our DAP model, the training resolution is $512 \times 1024$.

\noindent \textbf{Evaluation Datasets \& Metrics.} For evaluation, following prior work~\cite{cao2025panda}, we assess our method on two widely used indoor datasets, Matterport3D~\cite{chang2017matterport3d} and Stanford2D3D~\cite{armeni2017joint}, to evaluate its zero-shot performance.
For outdoor scenes, we use Deep360~\cite{li2022mode} as the test set, also under a zero-shot setting.
Since existing outdoor data are limited, we further introduce a new benchmark, DAP-Test, which consists of 1,343 high-quality outdoor images with accurate depth annotations. 
Following~\cite{Jiang2021UniFuseUF}, we evaluate depth estimation performance with metrics including Absolute Relative Error (\textit{AbsRel}), Root Mean Squared Error (\textit{RMSE}), and a percentage metrics $\delta_1$, where $i = 1.25$.

\noindent \textbf{Implementation Details.} All experiments are conducted on H20 GPUs. For model training, the learning rate is set to 5e-6 for the ViT backbone and 5e-5 for the decoders, using the Adam optimizer~\cite{kingma2014adam}. 
The loss weight $\lambda_\text{1}$, $\lambda_\text{2}$, $\lambda_\text{3}$, $\lambda_\text{4}$, $\lambda_\text{5}$ and $\lambda_\text{6}$ are set to 1.0, 0.4, 5.0, 2.0, 2.0 and 2.0, respectively.
Data augmentation includes color jittering, horizontal translation, and flipping following~\cite{cao2025panda}. Please refer to supplementary material for more details.

\begin{table}[t!]
\centering
\caption{Quantitative comparison on the proposed DAP-Test benchmark. 
Our DAP achieves the best performance across all metrics, demonstrating the effectiveness of large-scale data scaling and domain-consistent training.}
\setlength{\tabcolsep}{10pt} 
\begin{tabular}{lccc}
\toprule
Method & AbsRel($\downarrow$) & RMSE($\downarrow$) & $\delta_1$($\uparrow$) \\ 
\midrule
DAC \cite{DepthAnyCamera} & 0.3197 & \cellcolor{orange!25} 8.799 & 0.5193\\
Unik3D \cite{piccinelli2025unik3d} & \cellcolor{orange!25}0.2517 & 10.56 & \cellcolor{orange!25}0.6086 \\
DAP (Ours) & \cellcolor{red!25}0.0781 & \cellcolor{red!25}6.804 & \cellcolor{red!25}0.9370 \\
\bottomrule
\end{tabular}
\label{tab:test}
\end{table}

\begin{table*}[t]
\centering
\caption{\label{tab:ablation} 
Ablation study on the proposed components. The \colorbox{red!25}{\textbf{best}} and \colorbox{orange!25}{\textbf{second best}} performances are highlighted. 
}
\resizebox{\textwidth}{!}{
\begin{tabular}{ccccccccc}
\toprule
\multirow{2}{*}{\centering Distortion Map} & 
\multirow{2}{*}{\centering Geometry Loss} & 
\multirow{2}{*}{\centering Sharpness Loss} &
\multicolumn{3}{c}{Stanford2D3D (Indoor)} &
\multicolumn{3}{c}{Deep360 (Outdoor)} \\
\cmidrule(lr){4-6} \cmidrule(lr){7-9}
 &  &  & AbsRel $(\downarrow)$ & RMSE $(\downarrow)$ & $\delta_1$ $(\uparrow)$ & AbsRel $(\downarrow)$ & RMSE $(\downarrow)$ & $\delta_1$ $(\uparrow)$ \\
\midrule
\xmark & \xmark & \xmark & 0.1166 & 0.449 & 0.8409 & 0.0942 & 6.374 & 0.8396 \\
\checkmark & \xmark & \xmark & 0.1149 & 0.449 & 0.8440 & 0.0926 & \cellcolor{orange!25}6.297 & 0.8423 \\
\checkmark & \checkmark & \xmark & \cellcolor{orange!25}0.1112 & \cellcolor{orange!25}0.448 & \cellcolor{orange!25}0.8509 & \cellcolor{orange!25}0.0880 & 6.373 & \cellcolor{orange!25}0.8592 \\
\checkmark & \checkmark & \checkmark & \cellcolor{red!25}0.1084 & \cellcolor{red!25}0.442 & \cellcolor{red!25}0.8576 & \cellcolor{red!25}0.0862 & \cellcolor{red!25}6.212 & \cellcolor{red!25}0.8719 \\
\bottomrule
\end{tabular}}
\end{table*}

\begin{table}[t]
\centering
\caption{\label{tab:maskdepth}
Ablation on the range mask head (m). The \colorbox{red!25}{\textbf{best}} and 
\colorbox{orange!25}{\textbf{second best}} results are highlighted. 
}
\resizebox{\columnwidth}{!}{
\begin{tabular}{ccccc}
\toprule
\multirow{2}{*}{\centering Mask Depth} & 
\multicolumn{2}{c}{DAP-2M-Labeled} &
\multicolumn{2}{c}{Deep360} \\
\cmidrule(lr){2-3} \cmidrule(lr){4-5}
 & AbsRel $(\downarrow)$ & $\delta_1$ $(\uparrow)$ & AbsRel $(\downarrow)$ & $\delta_1$ $(\uparrow)$ \\
\midrule
10  & \cellcolor{orange!25}0.0801 & \cellcolor{orange!25}0.9315 & 0.0934 & 0.8493 \\
20  & 0.0823 & 0.9164 & 0.0873 & \cellcolor{orange!25}0.8668 \\
50  & 0.0864 & 0.9104 & \cellcolor{red!25}0.0843 & 0.8594 \\
\midrule
100 & \cellcolor{red!25}0.0793 & \cellcolor{red!25}0.9353 & \cellcolor{orange!25}0.0862 & \cellcolor{red!25}0.8719 \\
\ding{55} & 0.0832 & 0.9042 & 0.0938 & 0.8411\\
\bottomrule 
\end{tabular}}
\end{table}

\subsection{Qualitative and Quantitative Evaluation}

\noindent \textbf{Zero-shot Performance.} Table~\ref{table:zero} reports the zero-shot metric depth results on Stanford2D3D, Matterport3D, and Deep360. We compare our method with recent metric approaches and also list several scale-invariant methods only as reference. Note that scale-invariant models require ground-truth depth to obtain scale during evaluation, while our method predicts absolute metric scale directly from the input panorama without any post-alignment. Nevertheless, our results remain comparably good with them.

Across all three benchmarks, DAP consistently delivers the best performance. On Stanford2D3D and Matterport3D, it significantly lowers AbsRel while achieving markedly higher $\delta_1$, demonstrating strong generalization to unseen indoor scenes. On Deep360, DAP obtains the lowest AbsRel and RMSE and the highest $\delta_1$. These results show that DAP unifies indoor and outdoor panoramic depth estimation within a single foundation model and achieves state-of-the-art performance without any fine-tuning, demonstrating both the scalability of our data engine and the robustness of the three-stage pipeline.

\noindent \textbf{DAP-Test Performance.} Table~\ref{tab:test} presents the quantitative results on our proposed DAP-Test benchmark. 
Although this dataset is considered in-domain for our model, it serves as an essential evaluation of the effectiveness of large-scale data scaling and training strategies. 
Compared with previous state-of-the-art methods, DAP performs better than both DAC and Unik3D across all metrics. 
Specifically, DAP reduces the AbsRel from 0.2517 to 0.0781 and the RMSE from 10.563 to 6.804, while increasing $\delta_1$ from 0.6086 to 0.9307. 
These substantial improvements demonstrate the advantages of our data engine and unified framework in achieving accurate, metrically consistent, and robust depth estimation for panoramic outdoor scenes.

\noindent \textbf{Qualitative Comparison.} Figure~\ref{fig:qualitative} presents qualitative comparisons among DAC \cite{DepthAnyCamera}, Unik3D \cite{piccinelli2025unik3d}, and our DAP across diverse indoor and outdoor real-world scenes.  Compared with existing approaches, DAP generates sharper object boundaries and more coherent global geometry, particularly in regions with large depth discontinuities or complex scene layouts. For indoor environments, our model accurately preserves fine structural details such as furniture edges and wall boundaries, while Unik3D and DAC often exhibit over-smoothed or distorted depth transitions.  Notably, although Unik3D performs well on Deep360, its generalization to diverse real-world outdoor scenes is limited. In particular, existing methods fail to maintain stable depth structures in distant regions and sky areas, leading to distorted or collapsed predictions. In contrast, DAP maintains strong robustness and metric consistency even under such challenging conditions, achieving smooth and realistic representations of both near and far spatial structures. These visual results further confirm that our large-scale data training and range-aware design effectively improve both metric fidelity and visual realism in panoramic depth estimation.

Figure \ref{fig:qualitative2} presents a qualitative comparison on the Stanford2D3D dataset. Compared with existing methods, our approach more accurately reconstructs distant structures and maintains fine geometric details that other models tend to blur or oversmooth. 
The results also show that our method exhibits stronger scale-awareness, producing depth and illumination distributions that are visually closer to the ground truth. 
In particular, the distant wall and ceiling regions in our predictions retain consistent color gradients and structural integrity, demonstrating the effectiveness of our degradation-perceptive design in preserving global scale consistency while recovering high-frequency details.

\subsection{Ablation Studies}

To fairly evaluate the contribution of each component in our framework while minimizing the influence of external priors, we adopt DINOv3-Large as the encoder and fully fine-tune all variants. 
We conduct comprehensive ablation studies to analyze the effectiveness of key modules in DAP. Additional visual comparisons are provided in the supplementary material.

\noindent \textbf{Model Design.}
Table~\ref{tab:ablation} summarizes the ablation results. Starting from a baseline trained only with the standard $\mathcal{L}_{SILog}$ loss, we progressively add the distortion map, geometry-consistent losses, and sharpness-related losses.
The distortion map improves optimization stability under ERP distortion, while the geometry-centric losses ($\mathcal{L}_{normal}$ and $\mathcal{L}_{pts}$) further enhance structural consistency. Adding the sharpness-centric losses ($\mathcal{L}_{DF}$ and $\mathcal{L}_{grad}$) achieves the best performance, yielding the lowest AbsRel (0.1084/0.0862) and highest $\delta_1$ (0.8576/0.8719) on Stanford2D3D and Deep360.

\noindent \textbf{Range Mask Head.} We evaluate the plug-and-play range mask head using thresholds of 10 m, 20 m, 50 m, and 100 m (Table~\ref{tab:maskdepth}), with ground-truth depth truncated accordingly for fair comparison.
Smaller thresholds (10 m and 20 m) emphasize near-range geometry and achieve $\delta_1$ above 0.91 on DAP-2M-Labeled. The 100 m setting offers the best overall balance, achieving AbsRel of 0.0793/0.0862 and $\delta_1$ of 0.9353/0.8719 on DAP-2M-Labeled and Deep360.
Removing the mask notably degrades performance, confirming its effectiveness in filtering unreliable far-depth predictions and stabilizing training.
Overall, the range mask head provides a flexible and reliable mechanism for maintaining metric consistency across diverse scene scales.

\section{Conclusion}

In this paper, we propose a panoramic metric-depth foundation model DAP, which is built through large-scale data scaling and a unified three-stage training pipeline. By combining reliable pseudo-labeling, geometry-aware design, and a plug-and-play range mask head, 
DAP achieves strong zero-shot generalization and state-of-the-art performance across indoor–outdoor benchmarks, with particularly robust and stable metric predictions in diverse real-world outdoor environments.

\label{sec:conclusion}

{
    \small
    \bibliographystyle{ieeenat_fullname}
    \bibliography{main}

@String(PAMI = {IEEE Trans. Pattern Anal. Mach. Intell.})

@String(CVPR= {IEEE Conf. Comput. Vis. Pattern Recog.})

@String(ICCV= {Int. Conf. Comput. Vis.})

@String(ECCV= {Eur. Conf. Comput. Vis.})

@String(TIP  = {IEEE Trans. Image Process.})

@String(ICLR = {Int. Conf. Learn. Represent.})

@String(AAAI = {AAAI})

@String(PAMI  = {IEEE TPAMI})

@String(CVPR  = {CVPR})

@String(ICCV  = {ICCV})

@String(ECCV  = {ECCV})

@String(TIP   = {IEEE TIP})

@String(ICLR  = {ICLR})

@inproceedings{piccinelli2025unik3d,
  title={UniK3D: Universal Camera Monocular 3D Estimation},
  author={Piccinelli, Luigi and Sakaridis, Christos and Segu, Mattia and Yang, Yung-Hsu and Li, Siyuan and Abbeloos, Wim and Van Gool, Luc},
  booktitle={CVPR},
  year={2025}
}

@inproceedings{cao2025panda,
  title={PanDA: Towards Panoramic Depth Anything with Unlabeled Panoramas and Mobius Spatial Augmentation},
  author={Cao, Zidong and Zhu, Jinjing and Zhang, Weiming and Ai, Hao and Bai, Haotian and Zhao, Hengshuang and Wang, Lin},
  booktitle={CVPR},
  year={2025}
}

@inproceedings{DepthAnyCamera,
  title={Depth any camera: Zero-shot metric depth estimation from any camera},
  author={Guo, Yuliang and Garg, Sparsh and Miangoleh, S Mahdi H and Huang, Xinyu and Ren, Liu},
  booktitle={CVPR},
  year={2025}
}

@article{hu2024metric3d,
  title={Metric3d v2: A versatile monocular geometric foundation model for zero-shot metric depth and surface normal estimation},
  author={Hu, Mu and Yin, Wei and Zhang, Chi and Cai, Zhipeng and Long, Xiaoxiao and Chen, Hao and Wang, Kaixuan and Yu, Gang and Shen, Chunhua and Shen, Shaojie},
  journal={PAMI},
  year={2024}
}

@inproceedings{yin2023metric3d,
  title={Metric3d: Towards zero-shot metric 3d prediction from a single image},
  author={Yin, Wei and Zhang, Chi and Chen, Hao and Cai, Zhipeng and Yu, Gang and Wang, Kaixuan and Chen, Xiaozhi and Shen, Chunhua},
  booktitle={ICCV},
  year={2023}
}

@inproceedings{wang2025vggt,
  title={Vggt: Visual geometry grounded transformer},
  author={Wang, Jianyuan and Chen, Minghao and Karaev, Nikita and Vedaldi, Andrea and Rupprecht, Christian and Novotny, David},
  booktitle={CVPR},
  year={2025}
}

@article{feng2025dit360,
  title={DiT360: High-Fidelity Panoramic Image Generation via Hybrid Training},
  author={Feng, Haoran and Zhang, Dizhe and Li, Xiangtai and Du, Bo and Qi, Lu},
  journal={arXiv preprint arXiv:2510.11712},
  year={2025}
}

@article{wang2024qwen2,
  title={Qwen2-vl: Enhancing vision-language model's perception of the world at any resolution},
  author={Wang, Peng and Bai, Shuai and Tan, Sinan and Wang, Shijie and Fan, Zhihao and Bai, Jinze and Chen, Keqin and Liu, Xuejing and Wang, Jialin and Ge, Wenbin and others},
  journal={arXiv preprint arXiv:2409.12191},
  year={2024}
}

@article{wang2025moge2,
  title={MoGe-2: Accurate Monocular Geometry with Metric Scale and Sharp Details},
  author={Wang, Ruicheng and Xu, Sicheng and Dong, Yue and Deng, Yu and Xiang, Jianfeng and Lv, Zelong and Sun, Guangzhong and Tong, Xin and Yang, Jiaolong},
  journal={arXiv preprint arXiv:2507.02546},
  year={2025}
}

@inproceedings{wang2025moge,
  title={Moge: Unlocking accurate monocular geometry estimation for open-domain images with optimal training supervision},
  author={Wang, Ruicheng and Xu, Sicheng and Dai, Cassie and Xiang, Jianfeng and Deng, Yu and Tong, Xin and Yang, Jiaolong},
  booktitle={CVPR},
  year={2025}
}

@article{simeoni2025dinov3,
  title={Dinov3},
  author={Sim{\'e}oni, Oriane and Vo, Huy V and Seitzer, Maximilian and Baldassarre, Federico and Oquab, Maxime and Jose, Cijo and Khalidov, Vasil and Szafraniec, Marc and Yi, Seungeun and Ramamonjisoa, Micha{\"e}l and others},
  journal={arXiv preprint arXiv:2508.10104},
  year={2025}
}

@inproceedings{piccinelli2024unidepth,
  title={UniDepth: Universal monocular metric depth estimation},
  author={Piccinelli, Luigi and Yang, Yung-Hsu and Sakaridis, Christos and Segu, Mattia and Li, Siyuan and Van Gool, Luc and Yu, Fisher},
  booktitle={CVPR},
  year={2024}
}

@article{piccinelli2025unidepthv2,
  title={Unidepthv2: Universal monocular metric depth estimation made simpler},
  author={Piccinelli, Luigi and Sakaridis, Christos and Yang, Yung-Hsu and Segu, Mattia and Li, Siyuan and Abbeloos, Wim and Van Gool, Luc},
  journal={arXiv preprint arXiv:2502.20110},
  year={2025}
}

@inproceedings{rey2022360monodepth,
  title={360monodepth: High-resolution 360 monocular depth estimation},
  author={Rey, Manuel and Area, Mingze Yuan and Richardt, Christian},
  booktitle={CVPR},
  year={2022}
}

@inproceedings{yang2024depthanything,
  title={Depth anything: Unleashing the power of large-scale unlabeled data},
  author={Yang, Lihe and Kang, Bingyi and Huang, Zilong and Xu, Xiaogang and Feng, Jiashi and Zhao, Hengshuang},
  booktitle={CVPR},
  year={2024}
}

@article{yang2024depthanythingv2,
  title={Depth anything v2},
  author={Yang, Lihe and Kang, Bingyi and Huang, Zilong and Zhao, Zhen and Xu, Xiaogang and Feng, Jiashi and Zhao, Hengshuang},
  journal={NeurIPS},
  year={2024}
}

@article{he2024lotus,
  title={Lotus: Diffusion-based visual foundation model for high-quality dense prediction},
  author={He, Jing and Li, Haodong and Yin, Wei and Liang, Yixun and Li, Leheng and Zhou, Kaiqiang and Zhang, Hongbo and Liu, Bingbing and Chen, Ying-Cong},
  journal={arXiv preprint arXiv:2409.18124},
  year={2024}
}

@article{simulator,
  title={AirSim360: A Panoramic Simulation Platform within Drone View},
  author={Ge, Xian and Pan, Yuling and Zhang, Yuhang and Li, Xiang and Zhang, Weijun and Zhang, Dizhe and Wan, Zhaoliang and Lin, Xin and Zhang, Xiangkai and Liang, Juntao and others},
  journal={arXiv preprint arXiv:2512.02009},
  year={2025}
}

@article{lin2025one,
  title={One flight over the gap: A survey from perspective to panoramic vision},
  author={Lin, Xin and Ge, Xian and Zhang, Dizhe and Wan, Zhaoliang and Wang, Xianshun and Li, Xiangtai and Jiang, Wenjie and Du, Bo and Tao, Dacheng and Yang, Ming-Hsuan and others},
  journal={arXiv preprint arXiv:2509.04444},
  year={2025}
}

@inproceedings{lee2025hush,
  title={HUSH: Holistic Panoramic 3D Scene Understanding using Spherical Harmonics},
  author={Lee, Jongsung and Park, Harin and Lee, Byeong-Uk and Joo, Kyungdon},
  booktitle={CVPR},
  year={2025}
}

@article{wang2024depthanywhere,
  title={Depth anywhere: Enhancing 360 monocular depth estimation via perspective distillation and unlabeled data augmentation},
  author={Wang, Ning-Hsu Albert and Liu, Yu-Lun},
  journal={NeurIPS},
  year={2024}
}

@inproceedings{ai2024elite360d,
  title={Elite360d: Towards efficient 360 depth estimation via semantic-and distance-aware bi-projection fusion},
  author={Ai, Hao and Wang, Lin},
  booktitle={CVPR},
  year={2024}
}

@inproceedings{yun2023egformer,
  title={Egformer: Equirectangular geometry-biased transformer for 360 depth estimation},
  author={Yun, Ilwi and Shin, Chanyong and Lee, Hyunku and Lee, Hyuk-Jae and Rhee, Chae Eun},
  booktitle={ICCV},
  year={2023}
}

@inproceedings{ai2023hrdfuse,
  title={Hrdfuse: Monocular 360deg depth estimation by collaboratively learning holistic-with-regional depth distributions},
  author={Ai, Hao and Cao, Zidong and Cao, Yan-Pei and Shan, Ying and Wang, Lin},
  booktitle={CVPR},
  year={2023}
}

@inproceedings{shen2022panoformer,
  title={PanoFormer: panorama transformer for indoor 360$^\circ$ depth estimation},
  author={Shen, Zhijie and Lin, Chunyu and Liao, Kang and Nie, Lang and Zheng, Zishuo and Zhao, Yao},
  booktitle={ECCV},
  year={2022}
}

@inproceedings{zhuang2022acdnet,
  title={Acdnet: Adaptively combined dilated convolution for monocular panorama depth estimation},
  author={Zhuang, Chuanqing and Lu, Zhengda and Wang, Yiqun and Xiao, Jun and Wang, Ying},
  booktitle={AAAI},
  year={2022}
}

@inproceedings{tateno2018distortion,
  title={Distortion-aware convolutional filters for dense prediction in panoramic images},
  author={Tateno, Keisuke and Navab, Nassir and Tombari, Federico},
  booktitle={ECCV},
  year={2018}
}

@article{bai2024glpanodepth,
  title={GLPanoDepth: Global-to-local panoramic depth estimation},
  author={Bai, Jiayang and Qin, Haoyu and Lai, Shuichang and Guo, Jie and Guo, Yanwen},
  journal={TIP},
  year={2024}
}

@article{ai2024elite360m,
  title={Elite360M: Efficient 360 Multi-task Learning via Bi-projection Fusion and Cross-task Collaboration},
  author={Ai, Hao and Wang, Lin},
  journal={arXiv preprint arXiv:2408.09336},
  year={2024}
}

@inproceedings{benny2025sphereuformer,
  title={SphereUFormer: A U-Shaped Transformer for Spherical 360 Perception},
  author={Benny, Yaniv and Wolf, Lior},
  booktitle={CVPR},
  year={2025}
}

@article{shen2024revisiting,
  title={Revisiting 360 depth estimation with panogabor: A new fusion perspective},
  author={Shen, Zhijie and Lin, Chunyu and Nie, Lang and Liao, Kang and Lin, Weisi and Zhao, Yao},
  journal={arXiv preprint arXiv:2408.16227},
  year={2024}
}

@inproceedings{deng2025omnistereo,
  title={OmniStereo: Real-time Omnidireactional Depth Estimation with Multiview Fisheye Cameras},
  author={Deng, Jiaxi and Wang, Yushen and Meng, Haitao and Hou, Zuoxun and Chang, Yi and Chen, Gang},
  booktitle={CVPR},
  year={2025}
}

@inproceedings{peng2023high,
  title={High-resolution depth estimation for 360deg panoramas through perspective and panoramic depth images registration},
  author={Peng, Chi-Han and Zhang, Jiayao},
  booktitle={WACV},
  year={2023}
}

@article{jiang2021unifuse,
  title={Unifuse: Unidirectional fusion for 360 panorama depth estimation},
  author={Jiang, Hualie and Sheng, Zhe and Zhu, Siyu and Dong, Zilong and Huang, Rui},
  journal={RAL},
  year={2021}
}

@inproceedings{wang2020bifuse,
  title={Bifuse: Monocular 360 depth estimation via bi-projection fusion},
  author={Wang, Fu-En and Yeh, Yu-Hsuan and Sun, Min and Chiu, Wei-Chen and Tsai, Yi-Hsuan},
  booktitle={CVPR},
  year={2020}
}

@article{wang2024depth,
  title={Depth anywhere: Enhancing 360 monocular depth estimation via perspective distillation and unlabeled data augmentation},
  author={Wang, Ning-Hsu Albert and Liu, Yu-Lun},
  journal={NeurIPS},
  year={2024}
}

@article{da2,
  title={DA{\textsuperscript{2}}: Depth Anything in Any Direction},
  author={Li, Haodong and Zheng, Wangguangdong and He, Jing and Liu, Yuhao and Lin, Xin and Yang, Xin and Chen, Ying-Cong and Guo, Chunchao},
  journal={arXiv preprint arXiv:2509.26618},
  year={2025}
}

@inproceedings{mohadikar2025omnidiffusion,
  title={OmniDiffusion: Reformulating 360 Monocular Depth Estimation Using Semantic and Surface Normal Conditioned Diffusion},
  author={Mohadikar, Payal and Duan, Ye},
  booktitle={WACV},
  year={2025}
}

@inproceedings{bochkovskiydepth,
  title={Depth pro: Sharp monocular metric depth in less than a second},
  author = {Bochkovskiy, Aleksei and Delaunoy, Amaël and Germain, Hugo and Santos, Marcel and Zhou, Yichao and Richter, Stephan R and Koltun, Vladlen},
  booktitle={ICLR},
  year={2025}
}

@inproceedings{marigold,
  title={Repurposing diffusion-based image generators for monocular depth estimation},
  author={Ke, Bingxin and Obukhov, Anton and Huang, Shengyu and Metzger, Nando and Daudt, Rodrigo Caye and Schindler, Konrad},
  booktitle={CVPR},
  year={2024}
}

@article{wang2025jasmine,
  title={Jasmine: Harnessing Diffusion Prior for Self-supervised Depth Estimation},
  author={Wang, Jiyuan and Lin, Chunyu and Guan, Cheng and Nie, Lang and He, Jing and Li, Haodong and Liao, Kang and Zhao, Yao},
  journal={arXiv preprint arXiv:2503.15905},
  year={2025}
}

@misc{flux2024,
    author={{Black Forest Labs}},
    title={FLUX},
    year={2024},
    howpublished={\url{https://github.com/black-forest-labs/flux}},
}

@inproceedings{rombach2022high,
  title={High-resolution image synthesis with latent diffusion models},
  author={Rombach, Robin and Blattmann, Andreas and Lorenz, Dominik and Esser, Patrick and Ommer, Bj{\"o}rn},
  booktitle={CVPR},
  year={2022}
}

@article{li2025stereodiff,
  title={StereoDiff: Stereo-Diffusion Synergy for Video Depth Estimation},
  author={Li, Haodong and Wang, Chen and Lei, Jiahui and Daniilidis, Kostas and Liu, Lingjie},
  journal={arXiv preprint arXiv:2506.20756},
  year={2025}
}

@inproceedings{ho2020denoising,
  title={Denoising diffusion probabilistic models},
  author={Ho, Jonathan and Jain, Ajay and Abbeel, Pieter},
  booktitle={NeurIPS},
  year={2020}
}

@inproceedings{li2022omnifusion,
  title={Omnifusion: 360 monocular depth estimation via geometry-aware fusion},
  author={Li, Yuyan and Guo, Yuliang and Yan, Zhixin and Huang, Xinyu and Duan, Ye and Ren, Liu},
  booktitle={CVPR},
  year={2022}
}

@inproceedings{zheng2020structured3d,
  title={Structured3d: A large photo-realistic dataset for structured 3d modeling},
  author={Zheng, Jia and Zhang, Junfei and Li, Jing and Tang, Rui and Gao, Shenghua and Zhou, Zihan},
  booktitle={ECCV},
  year={2020}
}

@inproceedings{chang2017matterport3d,
  title={Matterport3D: Learning from RGB-D Data in Indoor Environments},
  author={Chang, Angel and Dai, Angela and Funkhouser, Thomas and Halber, Maciej and Niebner, Matthias and Savva, Manolis and Song, Shuran and Zeng, Andy and Zhang, Yinda},
  booktitle={3DV},
  year={2017}
}

@inproceedings{li2022mode,
  title={MODE: Multi-view omnidirectional depth estimation with 360 cameras},
  author={Li, Ming and Jin, Xueqian and Hu, Xuejiao and Dai, Jingzhao and Du, Sidan and Li, Yang},
  booktitle={European Conference on Computer Vision},
  pages={197--213},
  year={2022},
  organization={Springer}
}

@article{armeni2017joint,
  title={Joint 2d-3d-semantic data for indoor scene understanding},
  author={Armeni, Iro and Sax, Sasha and Zamir, Amir R and Savarese, Silvio},
  journal={arXiv preprint arXiv:1702.01105},
  year={2017}
}

@article{kingma2014adam,
  title={Adam: A method for stochastic optimization},
  author={Kingma, Diederik P and Ba, Jimmy},
  journal={arXiv preprint arXiv:1412.6980},
  year={2014}
}

@article{Jiang2021UniFuseUF,
  title={UniFuse: Unidirectional Fusion for 360$^\circ$ Panorama Depth Estimation},
  author={Hualie Jiang and Zhe Sheng and Siyu Zhu and Zilong Dong and Rui Huang},
  journal={IEEE Robotics and Automation Letters},
  year={2021},
  volume={6},
  pages={1519-1526}
}

@article{eigen2014depth,
  title={Depth map prediction from a single image using a multi-scale deep network},
  author={Eigen, David and Puhrsch, Christian and Fergus, Rob},
  journal={Advances in neural information processing systems},
  volume={27},
  year={2014}
}
}


\end{document}